\documentclass[conference,a4paper]{IEEEtran}

\usepackage[utf8]{inputenc}
\usepackage[T1]{fontenc}
\usepackage{amsmath,amssymb}
\usepackage{graphicx}
\usepackage{booktabs}
\usepackage{multirow}
\usepackage{url}
\usepackage{hyperref}
\usepackage{xcolor}
\usepackage{tabularx}
\usepackage{cite}
\usepackage{tikz}
\usetikzlibrary{arrows.meta,positioning,shapes.geometric}

\hypersetup{
  colorlinks=true,
  linkcolor=black,
  citecolor=black,
  urlcolor=blue
}

\begin{document}

\title{Koshur Diacritizer: A Byte-Level Sequence-to-Sequence Model for Kashmiri Diacritic Restoration}

\author{
\IEEEauthorblockN{Haq Nawaz Malik$^{*}$}
\IEEEauthorblockA{ORCID: \texttt{0009-0003-1994-7640}\\
HF: \texttt{@Omarrran} \\ X: \texttt{@HAQ\_NAWAZ\_MALIK}}
\and
\IEEEauthorblockN{Nahfid Nissar}
\IEEEauthorblockA{ORCID: \texttt{0009-0002-2805-4687}\\
HF: \texttt{@nafiboi} \\ X: \texttt{@NahfidN}}
\and
\IEEEauthorblockN{Faizan Iqbal}
\IEEEauthorblockA{ORCID: \texttt{0009-0002-8998-9347}\\
HF: \texttt{@faizaniqbal} \\ X: \texttt{@faizaniqbal\_\_52}}
\thanks{$^{*}$Main author. Project repository: \url{https://huggingface.co/Omarrran/koshur-diacritizer-byt5-small}}
}

\maketitle

\begin{abstract}
Kashmiri, an Indo-Aryan language written primarily in a modified Perso-Arabic script, relies heavily on diacritic marks to represent short vowels and other phonological distinctions. However, these marks are frequently omitted in digital text, creating ambiguity and reducing the effectiveness of downstream natural language processing (NLP) systems. Despite the importance of diacritic restoration for applications such as text-to-speech, grapheme-to-phoneme conversion, and machine translation, Kashmiri remains largely unexplored in this area.

In this work, we introduce Koshur Diacritizer, a ByT5-small model fine-tuned for Kashmiri  diacritic restoration, formulated as a sequence-to-sequence task that maps undiacritized Kashmiri text to its fully diacritized form. To support this task, we release a publicly available dataset of 23.7k aligned Kashmiri sentence pairs consisting of non-diacritic inputs and their corresponding diacritic targets. The proposed system combines script-aware normalization, alignment validation, and a skeleton-preserving inference mechanism to ensure that restored outputs retain the original base-letter sequence. By operating directly on UTF-8 bytes, the model naturally handles Unicode combining marks, orthographic variation, and script-specific characters without requiring a language-specific tokenizer.

Experimental evaluation on a held-out test set yields a Diacritic Error Rate on marked positions (DER\textsubscript{m}) of 0.2012 and a Word Error Rate (WER) of 0.2159. A native Kashmiri linguistic expert rated each evaluated test sample, yielding a mean reviewer-rated accuracy of 77.5\%. These results indicate that the model captures a substantial portion of Kashmiri diacritic patterns while highlighting opportunities for improvement on rare linguistic phenomena and length-related truncation errors. We publicly release the dataset, trained model, and evaluation artifacts to facilitate reproducible research and future advances in Kashmiri language technology.
\end{abstract}


%-----------------------------------------------------------------------
\section{Introduction}
\label{sec:intro}

Kashmiri, an Indo-Aryan language predominantly spoken in the Kashmir Valley, employs a modified Perso-Arabic script for its written form. Within this orthographic system, diacritic marks, which are combining characters positioned above or below base consonants, are fundamental for encoding short vowels, nasalization, and other crucial phonological distinctions. These marks are indispensable for disambiguation, as an undiacritized consonant skeleton can correspond to multiple lexical entries with distinct pronunciations and meanings.

Despite their linguistic significance, diacritics are frequently omitted in contemporary digital communication, news corpora, and web-scraped texts. This prevalent omission exacerbates the challenges already confronting Kashmiri as a low-resource language, characterized by limited datasets and inconsistent annotation standards. Consequently, downstream NLP systems, such as text-to-speech (TTS), grapheme-to-phoneme (G2P) converters, and machine translation models, receive degraded input when diacritics are absent.

Automatic diacritic restoration, the process of recovering missing marks from bare text, has been extensively investigated for languages like Arabic~\cite{zitouni2006maximum,abandah2015automatic,fadel2019arabic}, Hebrew~\cite{shmidman2020nakdan}, and Vietnamese~\cite{luu2012pointwise}. However, this task remains largely unexplored for Kashmiri. Traditional rule-based approaches necessitate comprehensive lexica and morphological analyzers, resources that are scarce for Kashmiri. Neural sequence-to-sequence models, conversely, offer a data-driven alternative capable of learning contextual restoration patterns directly from aligned examples.

This work introduces \textit{Koshur Diacritizer}, a byte-level encoder-decoder model specifically designed for Kashmiri diacritic restoration. The system fine-tunes \texttt{google/byt5-small}~\cite{xue2022byt5}, a language-agnostic byte-level Transformer, on a meticulously curated corpus of aligned diacritized and undiacritized Kashmiri sentence pairs derived from  complementary sources. The byte-level formulation offers a distinct advantage in this context: standard subword tokenizers, optimized for high-resource languages, often fragment or mishandle the combining marks and Kashmiri-specific characters central to the task. Operating directly on UTF-8 bytes mitigates this risk.

The main contributions of this work are:

\begin{enumerate}
\item We release a publicly available dataset of 23.7k aligned Kashmiri sentence pairs for diacritic restoration, consisting of undiacritized inputs and their fully diacritized counterparts, providing a new resource for Kashmiri NLP research.

\item To the best of our knowledge, we present the first dedicated neural model for Kashmiri diacritic restoration. We introduce \textit{Koshur Diacritizer}, a byte-level sequence-to-sequence model based on ByT5 that restores diacritics directly from undiacritized text.

\item We demonstrate that byte-level tokenization is an effective modeling strategy for Kashmiri Perso-Arabic script, as it naturally handles Unicode combining marks, orthographic variation, and script-specific characters without requiring a language-specific tokenizer. In addition, we introduce a skeleton-safe restoration framework that preserves the original base-letter sequence through alignment-aware preprocessing and inference-time verification, and we validate the system with both automatic metrics and native-expert human evaluation.
\end{enumerate}

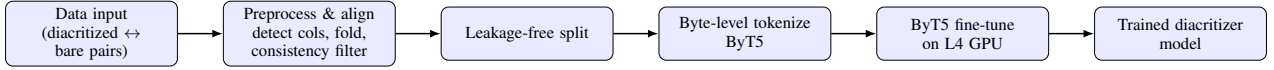
\begin{figure*}[!t]
\centering
\resizebox{0.92\textwidth}{!}{%
\begin{tikzpicture}[
node distance=0.85cm,
box/.style={draw, rounded corners, align=center, font=\small, minimum height=1.0cm, text width=3.0cm, fill=blue!8},
arrow/.style={-{Latex[length=2mm]}, thick}
]
\node[box] (A) {Data input\\(diacritized $\leftrightarrow$ bare pairs)};
\node[box, right=of A] (B) {Preprocess \& align\\detect cols, fold, consistency filter};
\node[box, right=of B] (C) {Leakage-free split};
\node[box, right=of C] (D) {Byte-level tokenize\\ByT5};
\node[box, right=of D] (E) {ByT5 fine-tune\\on L4 GPU};
\node[box, right=of E] (F) {Trained diacritizer\\model};

\draw[arrow] (A) -- (B);
\draw[arrow] (B) -- (C);
\draw[arrow] (C) -- (D);
\draw[arrow] (D) -- (E);
\draw[arrow] (E) -- (F);
\end{tikzpicture}%
}
\caption{Simplified training pipeline for the Koshur Diacritizer model.}
\label{fig:system-overview}
\end{figure*}

Figure~\ref{fig:system-overview} summarizes the core data preparation and model-training pipeline.

Section~\ref{sec:related} situates this work within the broader landscape of diacritic restoration and Kashmiri language technology.

%-----------------------------------------------------------------------
\section{Related Work}
\label{sec:related}

Building on the foundational problem of diacritic restoration, this section reviews prior research, highlighting both advancements in related languages and the specific gaps pertinent to Kashmiri.

\subsection{Diacritic Restoration for Arabic-Script Languages}

Diacritic restoration has gained significant attention, particularly for Modern Standard Arabic. Early statistical methods, such as maximum-entropy classifiers operating on character-level features, demonstrated initial success~\cite{zitouni2006maximum}. Subsequent research advanced through recurrent neural architectures, including bidirectional LSTMs with attention mechanisms~\cite{abandah2015automatic}, and more recently, Transformer-based models fine-tuned on extensive diacritized Arabic corpora~\cite{fadel2019arabic}. These systems benefit from the relative abundance of training data and the well-established morphological analysis available for Arabic.

However, directly transferring Arabic diacritization models to other Perso-Arabic script languages, such as Urdu, Persian, and Kashmiri, presents considerable challenges. Kashmiri, for instance, incorporates additional characters and diacritic marks not found in standard Arabic, and its morphological structure diverges significantly. Consequently, the literature lacks reports of prior neural diacritization systems specifically tailored for Kashmiri.

\subsection{Byte-Level Models for Low-Resource Languages}

The ByT5 family of models~\cite{xue2022byt5} represents a significant advancement by operating directly on raw UTF-8 bytes, thereby circumventing the need for language-specific tokenization. This characteristic is particularly advantageous for low-resource and morphologically rich languages, where conventional subword tokenizers, often trained on high-resource corpora, frequently produce suboptimal segmentations of unfamiliar scripts and combining characters. Byte-level models have consistently demonstrated competitive or superior performance in tasks involving noisy text and non-Latin scripts~\cite{xue2022byt5,clark2022canine}.

\subsection{Kashmiri in NLP world}

Kashmiri NLP remains in its nascent stages. Previous efforts have focused on corpus construction for Kashmiri literary texts~\cite{malik2025kslit}, transliteration between Perso-Arabic and Roman scripts, and preliminary explorations into machine translation. The current work expands this emerging ecosystem by addressing a fundamental text normalization task that directly supports the development of more advanced downstream applications.

This review underscores the critical need for a dedicated Kashmiri diacritic restoration system, a niche that this present work aims to fill by leveraging byte-level modeling for its unique orthographic characteristics.

%-----------------------------------------------------------------------
\section{Task Formulation}
\label{sec:task}

Given the identified gaps in Kashmiri diacritic restoration we  constrained text-generation problem in which the model must recover missing combining marks while preserving the original base-letter sequence.

The diacritic restoration task is defined as follows: let $x$ represent an input string in the Kashmiri Perso-Arabic script from which diacritic marks have been removed, and let $y$ denote the corresponding fully diacritized string. The primary objective is to learn a mapping $f: x \rightarrow y$ such that the base-letter skeleton of $y$ remains identical to $x$, while the combining marks in $y$ accurately represent the intended pronunciation. This formulation is distinct from machine translation, as the source and target share the same language, base-letter inventory, and largely preserve word order. The model's principal function is to insert or recover combining marks without altering the underlying consonant skeleton. Violations of this constraint, where the model rewrites base letters, are considered safety failures rather than mere accuracy errors.

To prevent the model from generating spurious base letters or altering the input's fundamental structure, a prediction $\hat{y}$ is formally considered \emph{skeleton-safe} if and only if:
\begin{equation}
\mathrm{strip}(\hat{y}, \mathcal{F}) = x
\label{eq:skeleton}
\end{equation}
where, $\mathrm{strip}(\cdot, \mathcal{F})$ is a function that removes all diacritic marks and applies a letter-fold map $\mathcal{F}$ to normalize script variants. This crucial constraint is enforced during inference through a post-generation guard, as detailed in Section~\ref{sec:model}. Realizing this formulation requires aligned training pairs that satisfy Equation~\ref{eq:skeleton}; the following section describes their construction.

%-----------------------------------------------------------------------
\section{Dataset and Preprocessing}
\label{sec:data}

The dataset is built through a robust pipeline whose every step records why it acted, so the corpus is reproducible and auditable.

\subsection{Source Corpora}

The training data for this study is drawn from a Kashmiri parallel diacritization dataset hosted on the Hugging Face Hub~\footnote{\url{https://huggingface.co/datasets/Omarrran/kashmiri_parallel_Diacratic_to_Non_diacratic_Text_dataset}}. The dataset contains 28,891 aligned sentence pairs, with each pair consisting of an undiacritized Kashmiri input and its corresponding fully diacritized target. The canonical dataset copy is used for preprocessing, orientation detection, cleaning, and alignment validation.

\subsection{Unicode Normalization and Letter Folding}

All text undergoes NFC normalization, tatweel (kashida) removal, and whitespace collapsing to ensure consistency. For accurate skeleton comparison, a learned letter-fold map $\mathcal{F}$ normalizes script variants. The expanded fold map employed in this work maps ten variant forms to their canonical counterparts (Table~\ref{tab:fold}). 

\begin{table}[t]
\centering
\caption{Learned letter-fold map for skeleton comparison. Entries marked with $\dagger$ were added cross checked by linguistic expert.}
\label{tab:fold}
\begin{tabular}{@{}ll@{}}
\toprule
\textbf{Source Variant} & \textbf{Canonical Form} \\
\midrule
Alef Wasla (U+0672) & Alef (U+0627) \\
Alef with Wavy Hamza Below (U+0673) & Alef (U+0627) \\
Alef with Madda Above (U+0622) & Alef (U+0627) \\
Alef with Hamza Above (U+0623) & Alef (U+0627) \\
Alef with Hamza Below (U+0625) & Alef (U+0627) \\
Alef Wasla (U+0671)$^\dagger$ & Alef (U+0627) \\
Waw with Hamza Above (U+0624) & Waw (U+0648) \\
Yeh with Hamza Above (U+0626) & Farsi Yeh (U+06CC) \\
Heh with Yeh Above (U+06C2)$^\dagger$ & Heh Goal (U+06C1) \\
Yeh Barree w/ Hamza (U+06D3)$^\dagger$ & Yeh Barree (U+06D2) \\
\bottomrule
\end{tabular}
\end{table}

\subsection{Alignment Filtering and Deduplication}

Training examples must strictly adhere to the skeleton-alignment constraint defined in Equation~\ref{eq:skeleton}. This means that after stripping diacritics and applying the fold map to the target string, the result must precisely match the input. Examples violating this condition, along with duplicates (identified by target content) and instances exceeding 200 characters, are systematically removed. Table~\ref{tab:filtering} summarizes the filtering outcome for the combined corpus, demonstrating the rigorous data quality control.

\begin{table}[t]
\centering
\caption{Dataset filtering statistics for the combined corpus.}
\label{tab:filtering}
\begin{tabular}{@{}lr@{}}
\toprule
\textbf{Criterion} & \textbf{Count} \\
\midrule
Combined raw input rows & 28,891 \\
Dropped by length ($>$200 chars) & 3,174 \\
Dropped as misaligned & 16 \\
Dropped as duplicate & 1,974 \\
\midrule
Retained pairs & 23,727 \\
Alignment survival rate & 82.13\% \\
\bottomrule
\end{tabular}
\end{table}

Of the 28,891 combined rows, 82.13\% survived the full pipeline, with the largest losses attributable to the 200-character length cap (3,174 rows) and deduplication (1,974 rows).

\subsection{Deterministic Split Strategy}

The corpus is partitioned using a deterministic content-hash strategy to ensure reproducibility across experiments. Each target string is hashed via $\mathrm{MD5}(\texttt{seed} \mathbin\Vert \texttt{target})$ with a fixed seed of 13. The resulting hash bucket then determines assignment to the test (5\%), validation (5\%), or training (90\%) splits. This approach guarantees invariance to data ordering and effectively minimizes data leakage risks. The final partition statistics are presented in Table~\ref{tab:splits}, where density denotes combining marks divided by the total character count of the diacritized string.

\begin{table}[t]
\centering
\caption{Split statistics after preprocessing. Density is combining marks over total characters of the diacritized string.}
\label{tab:splits}
\begin{tabular}{@{}lrrrr@{}}
\toprule
\textbf{Split} & \textbf{Rows} & \textbf{Mean chars} & \textbf{P95 chars} & \textbf{Density} \\
\midrule
Train & 21,295 & 113.6 & 191 & 0.1243 \\
Validation & 1,282 & 116.3 & 193 & 0.1248 \\
Test & 1,150 & 114.7 & 192 & 0.1245 \\
\bottomrule
\end{tabular}
\end{table}

%-----------------------------------------------------------------------
\section{Model Architecture and Training}
\label{sec:model}

The restoration mapping defined in Section~\ref{sec:task} is achieved through a byte-level sequence-to-sequence model.

\subsection{Architecture}

The system employs \texttt{google/byt5-small}~\cite{xue2022byt5} (approximately 300~million parameters) as its base architecture. This model is a byte-level variant of T5~\cite{raffel2020exploring}, where the tokenizer operates directly on UTF-8 byte sequences using a fixed vocabulary of 384 tokens. The architecture consists of 12 encoder layers and 4 decoder layers, featuring a hidden dimension of 1,472, 6 attention heads, and a feed-forward dimension of 3,584. This asymmetric, encoder-heavy design is characteristic of ByT5 models, reflecting the increased computational cost associated with processing longer byte sequences.

The byte-level formulation is critical for this task. Kashmiri Perso-Arabic characters frequently comprise a base letter followed by one or more combining marks, each encoded as distinct Unicode code points occupying multiple UTF-8 bytes. Standard subword tokenizers risk fragmenting these composite characters at arbitrary boundaries, potentially separating a base letter from its diacritics. Byte-level processing, conversely, preserves the full granularity of the character composition, thereby avoiding such issues.

\subsection{ Training}

The final model is trained on the filtered 23,727-pair corpus using \texttt{google/byt5-small} as the byte-level sequence-to-sequence backbone. Training runs for 10~epochs on a single NVIDIA L4 GPU and uses mixed-precision BF16 together with TF32 arithmetic where supported. The best checkpoint is selected according to the lowest validation DER\textsubscript{m}.
Table~\ref{tab:hyperparams} summarizes the training hyperparameters used for the final model.

\begin{table}[t]
\centering
\caption{Training hyperparameters (both stages).}
\label{tab:hyperparams}
\begin{tabular}{@{}lr@{}}
\toprule
\textbf{Parameter} & \textbf{Value} \\
\midrule
Epochs & 10 \\
Learning rate & $5 \times 10^{-4}$ \\
Weight decay & 0.01 \\
Warmup ratio & 0.05 \\
Scheduler & Cosine \\
Optimizer & AdamW \\
Per-device batch size & 16 \\
Gradient accumulation steps & 2 \\
Effective batch size & 32 \\
Max source length (bytes) & 256 \\
Max target length (bytes) & 256 \\
Generation beams & 1 (greedy) \\
Precision & BF16 \\
Seed & 42 \\
\bottomrule
\end{tabular}
\end{table}

\subsection{Inference with Skeleton Guard}

During inference, the model generates a diacritized output for a given undiacritized input. A post-generation skeleton guard then rigorously verifies that the prediction adheres to the constraint defined in Equation~\ref{eq:skeleton}. If the stripped prediction deviates from the input, the output is rejected, and the original input is returned unchanged. This mechanism proactively prevents the model from hallucinating alternate words during deployment, ensuring the integrity of the base text. For example, if the model generates a base letter not present in the input, the guard will reject the output. The evaluation metrics reported in Section~\ref{sec:results} reflect raw model predictions prior to the application of this skeleton guard, thereby representing the unfiltered generative behavior of the model rather than its guarded deployment behavior.

%-----------------------------------------------------------------------
\section{Evaluation Metrics}
\label{sec:metrics}

To comprehensively assess the model's performance, three complementary metrics are employed, each providing a distinct perspective on restoration accuracy.

\begin{enumerate}
\item \textbf{DER\textsubscript{m} (Diacritic Error Rate, marked positions):} This metric quantifies the fraction of incorrectly predicted diacritics exclusively among character positions that carry a diacritic in the reference text.

\item \textbf{DER\textsubscript{a} (Diacritic Error Rate, all positions):} This error rate is computed over all non-space base characters, encompassing both positions that should be marked and those that should remain unmarked.

\item \textbf{WER (Word Error Rate):} WER measures the word-level minimum edit distance between the fully diacritized reference and the model's prediction, normalized by the reference length.

\end{enumerate}

The DER computation involves segmenting each string into tuples of (base character, combining marks), aligning reference and hypothesis skeletons using \texttt{difflib.SequenceMatcher}, and then counting discrepancies in the combining-mark component for aligned base characters. Skeleton mismatches contribute the corresponding reference positions as errors. While DER quantifies mark-level accuracy, WER captures the impact on word integrity, and exact match provides a stringent measure of overall sentence correctness. These automatic metrics are complemented by native-expert human evaluation (Section~\ref{sec:human}).

%-----------------------------------------------------------------------
\section{Results}
\label{sec:results}

This section presents the empirical results of the  model,  human expert evaluation, and an analysis of training dynamics and the validation--test gap.

\subsection{Final Model Performance}

Table~\ref{tab:results} presents the final automatic evaluation metrics for the  model on both validation and test partitions. The test set comprises 1,150 sentences, containing 93,255 non-space character positions, of which 17,022 carry reference diacritics. The final model attains a Diacritic Error Rate on marked positions (DER\textsubscript{m}) of 0.2012, a Word Error Rate (WER) of 0.2159 held-out test set. As discussed in Section~\ref{sec:error}, the 256-byte generation budget truncates a portion of longer references, so these figures are best interpreted as upper bounds on the true mark-placement error.

\begin{table}[t]
\centering
\caption{Automatic evaluation metrics for the  model. Lower is better for all metrics }
\label{tab:results}
\begin{tabular}{@{}lcc@{}}
\toprule
\textbf{Metric} & \textbf{Validation} & \textbf{Test} \\
\midrule
Loss & 0.0611 & 0.0204\\
DER\textsubscript{m} & 0.1001 & 0.2012 \\
DER\textsubscript{a} & 0.0376 & 0.1469 \\
WER & 0.1231 & 0.2159 \\
Sentences & 1,282 & 1,150 \\
\bottomrule
\end{tabular}
\end{table}

\begin{table}[t]
\centering
\caption{Test-set metric comparison of the  model. Lower is better for all metrics .}
\label{tab:stage}
\begin{tabular}{@{}lcc@{}}
\toprule
\textbf{Metric} & \textbf{Final (23,727)} \\
\midrule
DER\textsubscript{m} & 0.2012 \\
DER\textsubscript{a} & 0.1469 \\
WER & 0.2159 \\
\bottomrule
\end{tabular}
\end{table}

\subsection{Human Expert Evaluation}
\label{sec:human}

Automatic metrics such as DER and WER penalize every mark mismatch equally, yet not all diacritic deviations are perceptually or linguistically significant. To complement the automatic evaluation, a native Kashmiri reviewer with linguistic expertise independently assessed each evaluated test sample. For every sample the reviewer assigned a correctness rating on a {0--100\% per sample} scale reflecting diacritic correctness and pronunciation fidelity, without access to the automatic scores. Across the {60} rated samples, the mean reviewer-rated accuracy was \textbf{77.5\%} (Table~\ref{tab:human}).

Notably, the human rating exceeds what the test DER\textsubscript{m} of 0.2012 might suggest. This indicates that a meaningful share of mark-level ``errors'' are perceptually acceptable variants or arise from output truncation (Section~\ref{sec:error}) rather than incorrect vowel choices, reinforcing the need for truncation-aware and severity-weighted evaluation.

\begin{table}[t]
\centering
\caption{Human expert evaluation summary.}
\label{tab:human}
\begin{tabular}{@{}lr@{}}
\toprule
\textbf{Aspect} & \textbf{Value} \\
\midrule
Reviewers & 2 (native, linguistic expert) \\
Samples rated & {60} \\
Rating scale & { 0--100\% per sample} \\
Mean rated accuracy & 77.5\% \\
\bottomrule
\end{tabular}
\end{table}

\subsection{Training Dynamics}

During the last train run, the training stages, loss consistently decreased from 0.1435 at the first logged step to 0.0217 at the final logged step, achieving a minimum of 0.0202. The best validation DER\textsubscript{m} (0.1001) was recorded at step~5,989 (epoch~9.0), with a corresponding validation loss of 0.0611.

\subsection{Validation--Test Gap}

A notable discrepancy persists between validation and test performance, with the test DER\textsubscript{m} approximately twice the validation value. This gap can be attributed to several contributing factors, including the moderate partition sizes (1,282 validation and 1,150 test sentences), byte-length truncation affecting longer outputs, and the inherently strict nature of exact-match scoring. This suggests that the model's performance on unseen, more diverse data is less optimistic than indicated by validation metrics.

%-----------------------------------------------------------------------
\section{Error Analysis}
\label{sec:error}

This section examines diacritic confusion patterns, truncation effects, and broader interpretations of the model's behavior, which together point to specific architectural and data remedies discussed in Section~\ref{sec:discussion}.

\subsection{Diacritic Confusion Patterns}

Table~\ref{tab:confusion} enumerates the ten most frequent confusion pairs observed in the test predictions, revealing two dominant error modes. The first is spurious mark insertion, notably bare~$\rightarrow$~kasra (331 occurrences) and bare~$\rightarrow$~fatha (208 occurrences). The second is mark omission, exemplified by kasra~$\rightarrow$~bare (307 occurrences) and hamza below~$\rightarrow$~bare (228 occurrences). The increased prominence of hamza-below confusions in the expanded model likely reflects the greater prevalence of hamza-bearing forms within the dataset. Inter-mark confusions, such as hamza below~$\leftrightarrow$~kasra (94 and 61 occurrences in opposite directions) and kasra~$\rightarrow$~damma (65 occurrences), suggest residual ambiguity in vowel identity, indicating the model struggles with fine-grained diacritic distinctions.

\begin{table}[t]
\centering
\caption{Top diacritic confusions on the test set. ``Bare'' denotes the absence of a combining mark.}
\label{tab:confusion}
\begin{tabular}{@{}llr@{}}
\toprule
\textbf{Reference} & \textbf{Predicted} & \textbf{Count} \\
\midrule
Bare & Kasra & 331 \\
Kasra & Bare & 307 \\
Hamza Below & Bare & 228 \\
Bare & Fatha & 208 \\
Bare & Hamza Below & 206 \\
Fatha & Bare & 158 \\
Damma & Bare & 156 \\
Bare & Damma & 155 \\
Hamza Below & Kasra & 94 \\
Bare & Inv.\ Small V Above & 73 \\
\bottomrule
\end{tabular}
\end{table}

\subsection{Truncation Effects}

A substantial proportion of test errors stem not from incorrect diacritic selection but from output truncation. Kashmiri Perso-Arabic characters in the U+0600--U+06FF block each occupy two UTF-8 bytes, so the 256-byte generation cap corresponds to only about 126 characters. Given a test-set P95 length of 192 characters (Table~\ref{tab:splits}), a non-trivial fraction of references exceed the cap and are necessarily truncated; the same cap applies to the training targets, so the model is in part trained to emit truncated output. Consistent with this, among the 60 saved sample predictions, 41 exhibited reference byte counts exceeding the 256-byte cap, 44 predictions terminated near this cap, and 15 had predicted character counts below 80\% of the reference length. Because skeleton mismatches in truncated regions are charged as diacritic errors, the DER and WER reported in Section~\ref{sec:results} should be interpreted as upper bounds on the true mark-placement error. A full-corpus quantification of the truncated fraction and a length-stratified DER are deferred to future work (Section~\ref{sec:discussion}).

\subsection{Interpretation}

The observed confusion profile is characteristic of a low-resource restoration model trained without explicit access to a lexicon or morphological analyzer. In many contexts, the bare consonant skeleton alone does not uniquely determine the diacritized form, particularly for function words and inflectional affixes where vowel patterns vary significantly by grammatical context. Incorporating lexicon constraints, morphological features, or longer contextual windows could effectively mitigate these ambiguities and improve restoration accuracy.

%-----------------------------------------------------------------------
\section{Discussion}
\label{sec:discussion}

This section discusses the strengths of the current approach, acknowledges its limitations, and proposes concrete recommendations for future work, building upon the error analysis presented previously.

\subsection{Strengths of the Current Approach}

The proposed approach offers several notable strengths. First, it introduces a publicly available dataset of 23.7k aligned Kashmiri sentence pairs, addressing a major resource gap for Kashmiri NLP. Second, the byte-level ByT5 formulation naturally handles Unicode combining marks, orthographic variation, and script-specific characters without requiring a language-specific tokenizer, making it well suited to Kashmiri Perso-Arabic text. Third, the skeleton-safe restoration framework preserves the original base-letter sequence through alignment-aware preprocessing and inference-time verification, improving reliability and reducing unintended word modifications. Finally, the public release of the dataset, model, and evaluation artifacts provides a reproducible baseline and establishes a foundation for future research in Kashmiri diacritic restoration and related language technologies.

\subsection{Limitations}

Despite its contributions, this work has several limitations. First, the study does not include comparisons against rule-based systems, lexicon-driven approaches, or alternative neural architectures, making it difficult to contextualize performance relative to other restoration strategies. Second, the current model operates with a fixed maximum sequence length, which can lead to truncation on longer sentences and negatively affect restoration accuracy. Third, human evaluation was conducted by a single native Kashmiri linguistic expert; broader assessment involving multiple annotators would provide a more comprehensive measure of output quality and inter-annotator agreement. Finally, although the released dataset is one of the largest resources currently available for Kashmiri diacritic restoration, it may not fully capture the linguistic diversity of different domains, writing styles, and dialectal variations. Consequently, the model's generalization to unseen registers of Kashmiri remains an open question.

\subsection{Future Work and Recommendations}

Several directions may further improve Kashmiri diacritic restoration. First, future studies should evaluate larger byte-level architectures and compare them with character-level and multilingual transformer models to better understand the trade-offs between model capacity and restoration accuracy. Second, increasing the maximum sequence length or adopting more efficient long-context strategies may help mitigate truncation-related errors observed in longer sentences. Third, incorporating lexical resources, morphological information, or constrained decoding techniques could improve the restoration of ambiguous diacritic patterns that cannot be reliably inferred from local context alone.

From a resource perspective, expanding the dataset to include additional domains, dialectal variation, and contemporary text sources would improve model robustness and generalization. Future evaluations should also involve multiple native Kashmiri experts to obtain more comprehensive human assessments and measure inter-annotator agreement. Finally, the released dataset and model provide a foundation for exploring related applications, including grapheme-to-phoneme conversion, text-to-speech synthesis, speech recognition, machine translation, and other text normalization tasks for Kashmiri.

%-----------------------------------------------------------------------

%-----------------------------------------------------------------------
\section{Conclusion}
\label{sec:conclusion}

This work presents Koshur Diacritizer, a byte-level sequence-to-sequence system for automatic Kashmiri diacritic restoration. Addressing the scarcity of resources for Kashmiri NLP, we release a publicly available dataset of 23.7k aligned sentence pairs consisting of undiacritized Kashmiri text and their fully diacritized counterparts. To support reliable restoration, we develop a skeleton-safe framework that combines script-aware normalization, alignment validation, and inference-time verification to preserve the original base-letter structure of the input text.

Built upon ByT5-small, the proposed model formulates diacritic restoration as a conditional text generation task, directly mapping non-diacritic Kashmiri text to its diacritized form. Our experiments demonstrate that byte-level modeling is a practical and effective approach for Kashmiri Perso-Arabic script, as it naturally handles Unicode combining marks, orthographic variation, and script-specific characters without requiring a language-specific tokenizer. On a held-out test set, the model achieves a DER\textsubscript{m} of 0.2012 and a WER of 0.2159. In addition, human evaluation by a Kashmiri linguistic expert yields an average reviewer-rated accuracy of approximately 77.5\%, indicating that the model successfully captures a substantial portion of Kashmiri diacritic patterns despite the challenges of limited resources and orthographic ambiguity.

While the current system demonstrates promising performance, several opportunities remain for improvement. Future work should investigate larger byte-level models, longer context windows to mitigate truncation effects, lexicon- and morphology-aware decoding strategies, and broader human evaluation across multiple annotators. We hope that the resources and findings presented in this work will encourage further research on Kashmiri and contribute to the development of NLP technologies for low-resource languages.

\section{Resources and Availability}
\label{sec:resources}

To promote transparency and reproducibility, all resources associated with this work are publicly available:

\begin{itemize}
\item \textbf{Model:} \url{https://huggingface.co/Omarrran/koshur-diacritizer-byt5-small}
\item \textbf{Dataset:} \url{https://huggingface.co/datasets/Omarrran/kashmiri_parallel_Diacratic_to_Non_diacratic_Text_dataset}
\item \textbf{Source Code:} \url{https://huggingface.co/Omarrran/koshur-diacritizer-byt5-small/tree/main/Source%20Code}
\end{itemize}

\clearpage
\onecolumn
\appendices

\section{Sample Outputs}
\label{app:samples}

\begin{figure}[htbp]
\centering
\includegraphics[width=0.96\textwidth]{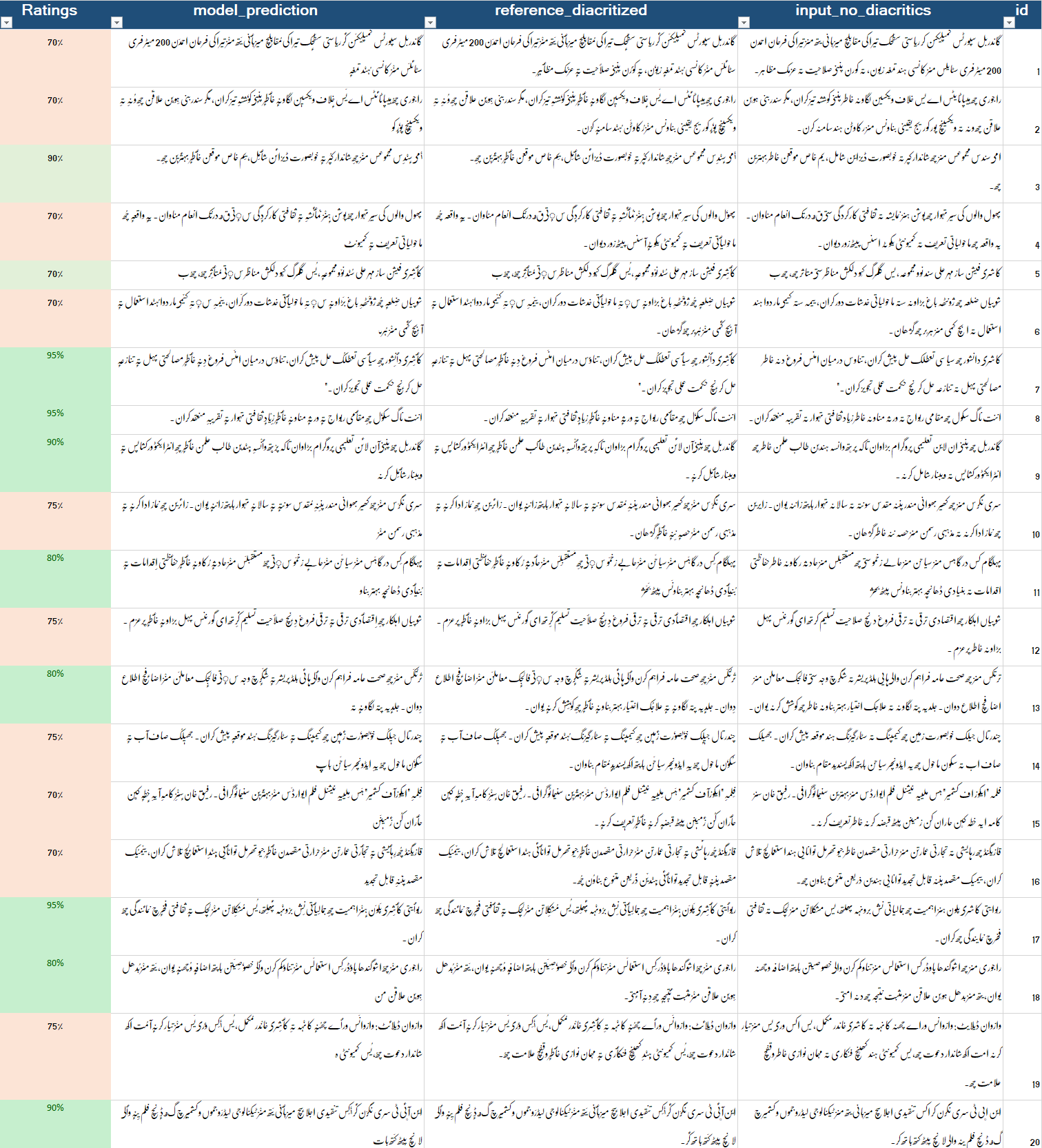}
\caption{Representative sample outputs from the Koshur Diacritizer system, part 1.}
\label{fig:sample-outputs-1}
\end{figure}

\begin{figure}[htbp]
\centering
\includegraphics[width=0.96\textwidth]{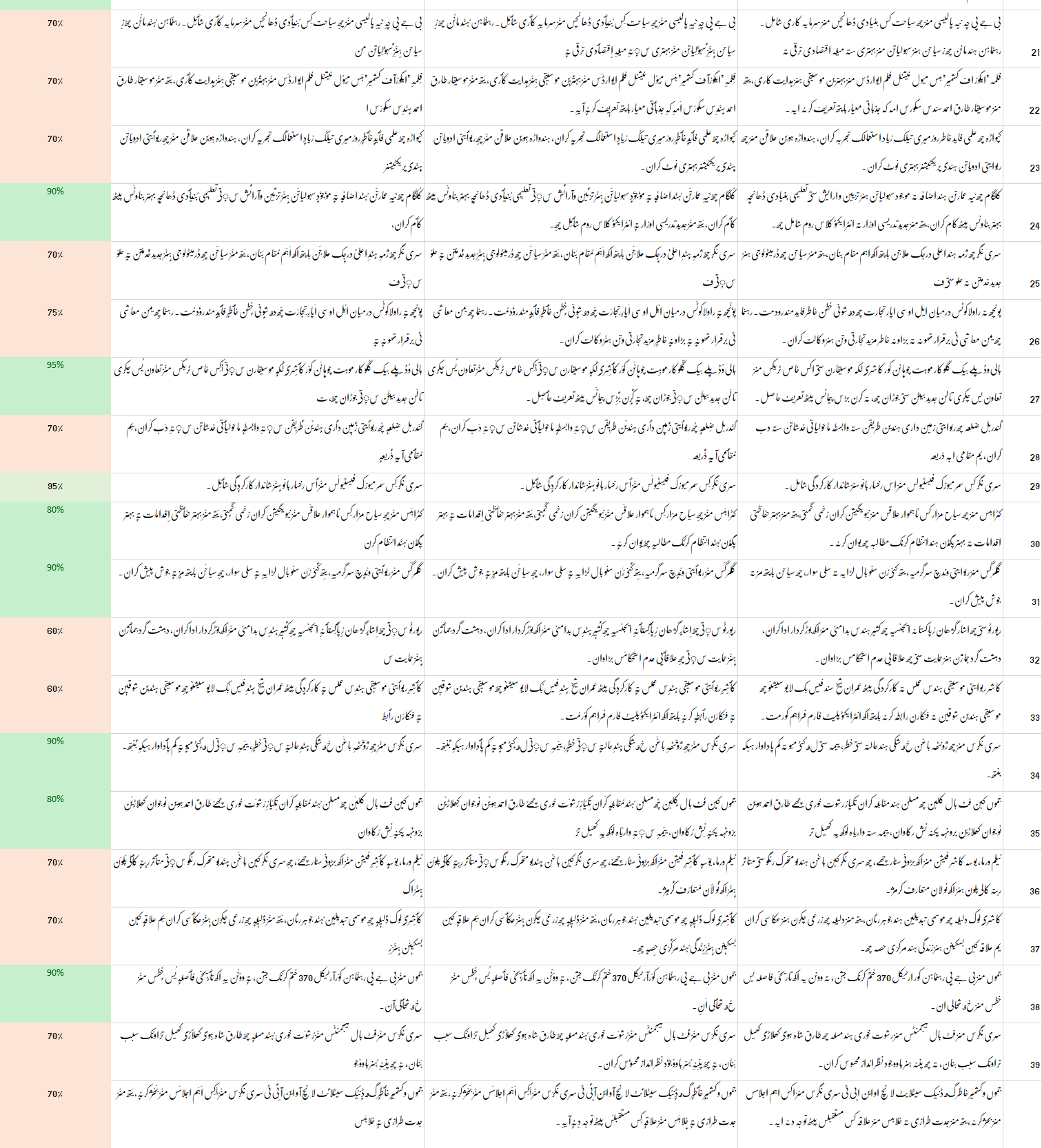}
\caption{Representative sample outputs from the Koshur Diacritizer system, part 2.}
\label{fig:sample-outputs-2}
\end{figure}

\begin{figure}[htbp]
\centering
\includegraphics[width=0.96\textwidth]{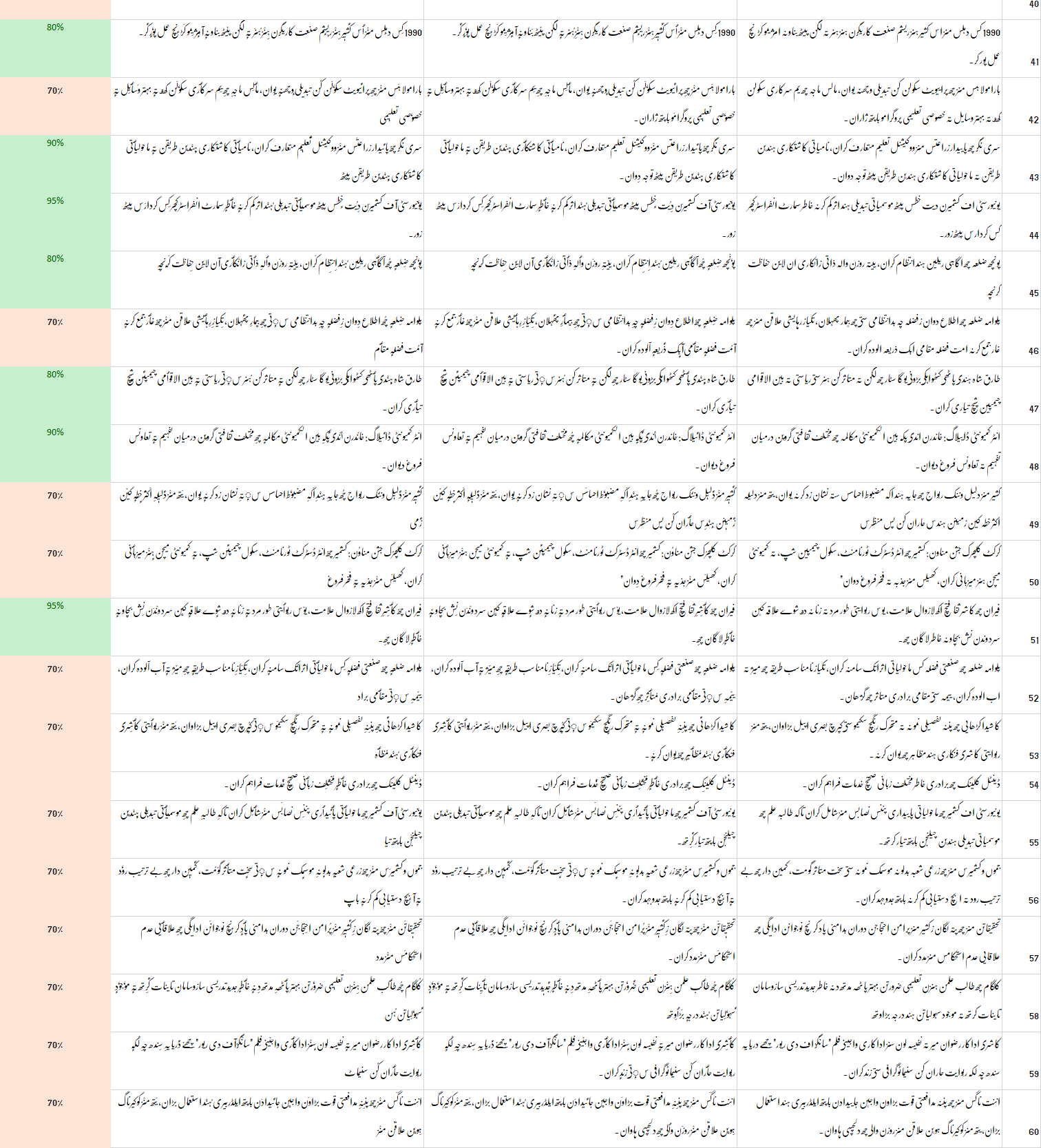}
\caption{Representative sample outputs from the Koshur Diacritizer system, part 3.}
\label{fig:sample-outputs-3}
\end{figure}

%-----------------------------------------------------------------------
\bibliographystyle{IEEEtran}

\end{document}